# Tool wear monitoring using an online, automatic and low cost system based on local texture

María Teresa García-Ordás[a,*], Enrique Alegre Gutiérrez[a], Rocío Alaiz-Rodríguez[a], Víctor González-Castro[a]

[a]*Department of Electrical, Systems and Automatic Engineering, Universidad of León. Campus de Vegazana s/n, 24071 León*

**Abstract**

In this work we propose a new online, low cost and fast approach based on computer vision and machine learning to determine whether cutting tools used in edge profile milling processes are serviceable or disposable based on their wear level. We created a new dataset of 254 images of edge profile cutting heads which is, to the best of our knowledge, the first publicly available dataset with enough quality for this purpose. All the inserts were segmented and their cutting edges were cropped, obtaining 577 images of cutting edges: 301 functional and 276 disposable. The proposed method is based on (1) dividing the cutting edge image in different regions, called Wear Patches (WP), (2) characterising each one as worn or serviceable using texture descriptors based on different variants of Local Binary Patterns (LBP) and (3) determine, based on the state of these WP, if the cutting edge (and, therefore, the tool) is serviceable or disposable. We proposed and assessed five different patch division configurations. The individual WP were classified by a Support Vector Machine (SVM) with an intersection kernel. The best patch division configuration and texture descriptor for the WP achieves an accuracy of 90.26% in the detection of the disposable cutting edges. These results show a very promising opportunity for automatic wear monitoring in edge profile milling processes.

*Corresponding Author

*Email addresses:* mgaro@unileon.es (María Teresa García-Ordás), ealeg@unileon.es (Enrique Alegre Gutiérrez), rocio.alaiz@unileon.es (Rocío Alaiz-Rodríguez), victor.gonzalez@unileon.es (Víctor González-Castro)

## 1. Introduction

The quality of the machined parts in milling, turning or drilling operations largely depends on the state of the cutting inserts. There are factors like abrasion, corrosion or fatigue that influence tool wear [1, 2]. Thus, tool wear monitoring becomes crucial in machining processes in order to find the optimal tool replacement time. This plays an important role not only because of the cost of the cutting tools themselves, but also for the indirect costs due to the unproductive time needed to carry out the tool replacement. Therefore, optimizing tool replacement operations significantly improves efficiency and competitiveness of the manufacturing systems.

The most widely studied techniques to assess the status of the cutting tools are based on monitoring signals that have some correlation with the level of wear. Such techniques are known as indirect methods. Thus, some works propose the use of force signals to measure wear in real time [3, 4, 5]. Another example is the work developed by Drazen et al. [6] in which they examine the influence of three cutting parameters on surface roughness, tool wear and cutting force components in face milling as part of an off-line process control. Other works are based on vibrations such as the one developed by Painuli et al. [7] in which descriptive statistical features from vibration signals are used in an online cutting tool condition monitoring system, or the work by Wafaa et al. [8] in which the vibratory signatures produced during a turning process were measured by using a three-axis accelerometer to monitor the wear on the tool. Acoustic emissions have been also shown to be sensitive to changes in cutting process conditions [9]. However, indirect methods present an important drawback: all these signals can be seriously affected by the inherent noise in industrial environments, which reduces their performance.

Recent advances in digital image processing have led to proposals for tool condition monitoring using machine vision, which are gaining importance day by day. In this case, tool wear is measured directly, achieving higher levels of precision and reliability than that of indirect methods. These computer vision approaches are mainly based on the wear shape contour, the wear shape properties, the texture of the wear region or combinations of some of them [10, 11, 12, 13, 14, 15, 16, 17, 18].

Antić et al. [14] apply a texture filter bank over the image formed by the Short Term Discrete Fourier Transform (STDFT) spectra of vibration sensors to get descriptors for tool wear monitoring. In [15] and [17], Miko-lajczyk et al. used unsupervised classification by means of Artificial Neural Networks for segmenting the tool wear region and thereafter, use it to predict the tool wear life. D'Adonna et al. use Artificial Neural Networks and DNA-based computing for predicting, based on information extracted from preprocessed images of tool wear images, the degree of wear [18]. Some works attempt to describe the wear taking into account the wear shape contour [10, 19]. Both methods are based on the ZMEG (Zernike Moment Edge Gradient) shape descriptor [20] obtaining promising results in the tool wear monitoring field. However, the use of just shape information may be quite limited to describe the wear since there are many other factors that characterize its level. In [12], the wear region is described using nine geometrical descriptors. This study shows that eccentricity, extent and solidity are the top three most informative features regarding wear level categorization. However, most methods based on geometric features have their limitations when applied to real environments, because the correct extraction of these features depends on a precise segmentation which becomes a complex stage due to the machining operation and other factors such as illumination conditions. In [11], some image processing techniques are developed to quantify two wear mechanisms (abrasion and micro-pitting) in polymers. They are based on local and global thresholding segmentation with different possible corrections and addresses the labelling processes as the mean opinion scores given by several experts, what is an interesting approach for this challenging task. This study paves the way towards an automated system to identify different wear mechanisms in this field.

Nowadays, the analysis of the wear region texture is the most widely studied approach. Thus, Datta et al. [21] rely on texture analysis and the extracted features were correlated with measured tool flank wear of turned surfaces. In [22], image analysis of surface textures produced during machining operations are used as indicators for predicting the condition (e.g., the wear) of the cutting tool. A computer vision that uses the grey level co-occurrence matrix (GLCM) for characterising surface roughness under different feed rates in face turning operations is proposed in [23]. Dutta et al. implemented an online acquisition system of machined surface images [24] which were subsequently analysed using an improvised GLCM technique with appropriate pixel pair spacing or offset parameter in turning processes. Later, the same authors proposed to use the discrete wavelet transform on turned surface images [13] and also texture analysis and support vector regression [25]. None of these works, however, deal with edge profile milling processes. In [26] a reliable machine vision system to automatically detect inserts and determine if they are broken is presented. Unlike the machining operations studied in the literature, they are dealing with edge milling head tools for aggressive machining of thick plates (up to 12 centimetres) in a single pass.

In this paper, we propose a new method based on image texture analysis for tool wear monitoring in an edge profile milling machine that can be embedded in a portable system. We consider it an *online* method because this evaluation is carried out with no external intervention to extract the inserts from the tool head. Our approach is based on obtaining local texture features individually from several regions extracted from the zone of the tool where the wear tends to appear according to the experts' knowledge and corroborated by the hundreds of inserts images captured. We will refer to these regions as Wear Patches (WP). This approach presents two main advantages: firstly, establishing WP with different sizes and orientations allows to detect small - but important- worn areas that otherwise (i.e. using methods that extract a single feature vector from the whole image) would have been overlooked. Secondly, it avoids the segmentation stage, which saves time and computational resources, making feasible a low cost portable implementation. Additionally, since each WP classification is addressed individually using supervised learning techniques based on kernels (i.e., SVM), the monitoring system is able to provide an estimation of the tool wear percentage by aggregating the individual results. This method can be successfully implemented in a small single-board computer, e.g. a Raspberry Pi. We consider that it is a *low-cost* system because, taking into account the components of the implemented system (i.e. the single-board computer, digital camera and LED bars that compose the illumination system), the cost may be in the range of €1500 -€2100.

The rest of this paper is organized as follows: Section 2 introduces the full process used for tool wear monitoring. Section 3 provides a description about the materials and methods that have been used in our experiments. Specifically, an explanation of the image acquisition process is detailed in Section 3.1, a study about the way to extract the patches of the images is presented in Section 3.2 and a description of the texture methods employed (i.e. LBP (Local Binary Pattern), ALBP (Adaptive Local Binary Pattern), CLBP (Completed Local Binary Pattern) and LBPV (LBP Variance)) is provided in Section 3.3. A new image dataset with cutting edges of an edge profile milling head is presented in Section 4 along with the experimental results and finally, the conclusions are listed in Section 5.

## 2. Proposed monitoring system and related work

In this work we propose a monitoring systems to assess the tool wear state in edge profile milling processes using computer vision and machine learning techniques. The outline of the whole methodology is shown in Figure 1 where the training and operational stages are shown. Next, we briefly describe the whole process and in the following sections, we elaborate on each one.

First of all, a portable system is used to take gray scale images of the tool head. Afterwards, the cutting edges are automatically detected and extracted. Then, each cutting edge is divided in several wear patches (WP) and each one is characterised using texture descriptors based on LBP. Finally, each WP is classified using a Support Vector Machine (SVM), and these classifications are used to make a final decision about the wear state of the tool cutting edge.

The background has a significant influence on the segmentation results, as Zhu and Yu pointed out in [27], where the wear region of the micro-milling tool image is segmented, as its size is a clear indicator of the tool wear condition. First, they used Morphological Component Analysis (MCA) to decompose the original image into target tool image, background and noise. After that, they applied a region growing algorithm on the target tool image only, thus, avoiding the side effect of the noise and background. In contrast, our approach assesses the wear condition by classifying the cutting edge using texture descriptors instead of using the area of the wear region. Additionally, the extraction of the cutting edge uses prior knowledge about the geometry of the inserts. Our proposal detects (1) the screw located in the middle of the inserts using the Circular Hough Transform [28] and then (2) the vertical and horizontal edges of the cutting tool using the method proposed by Canny [29] and the Standard Hough Transform [30] what makes the method very robust against changes in the background of the acquired images.

Other works have used description of different patches extracted from an image for different object recognition tasks such as 3D face recognition [31] or pedestrian detection [32, 33] merging the features extracted from each block. Instead, we describe and classify each patch individually, so the system is able to provide an estimation of the degree of tool wear by aggregating these individual classifications. Note that this gives robustness to the final decision, as a possible misclassification of a few single blocks would not affect the final tool wear estimation system as much as if method misclassified the whole image. Furthermore, using such multi-block approach, our system is able to detect small regions of high wear in the image which might have been ignored otherwise. It is important to highlight that our system does not require any segmentation of the tool wear region, which is a very time consuming step and, moreover, might not always be accurate enough.

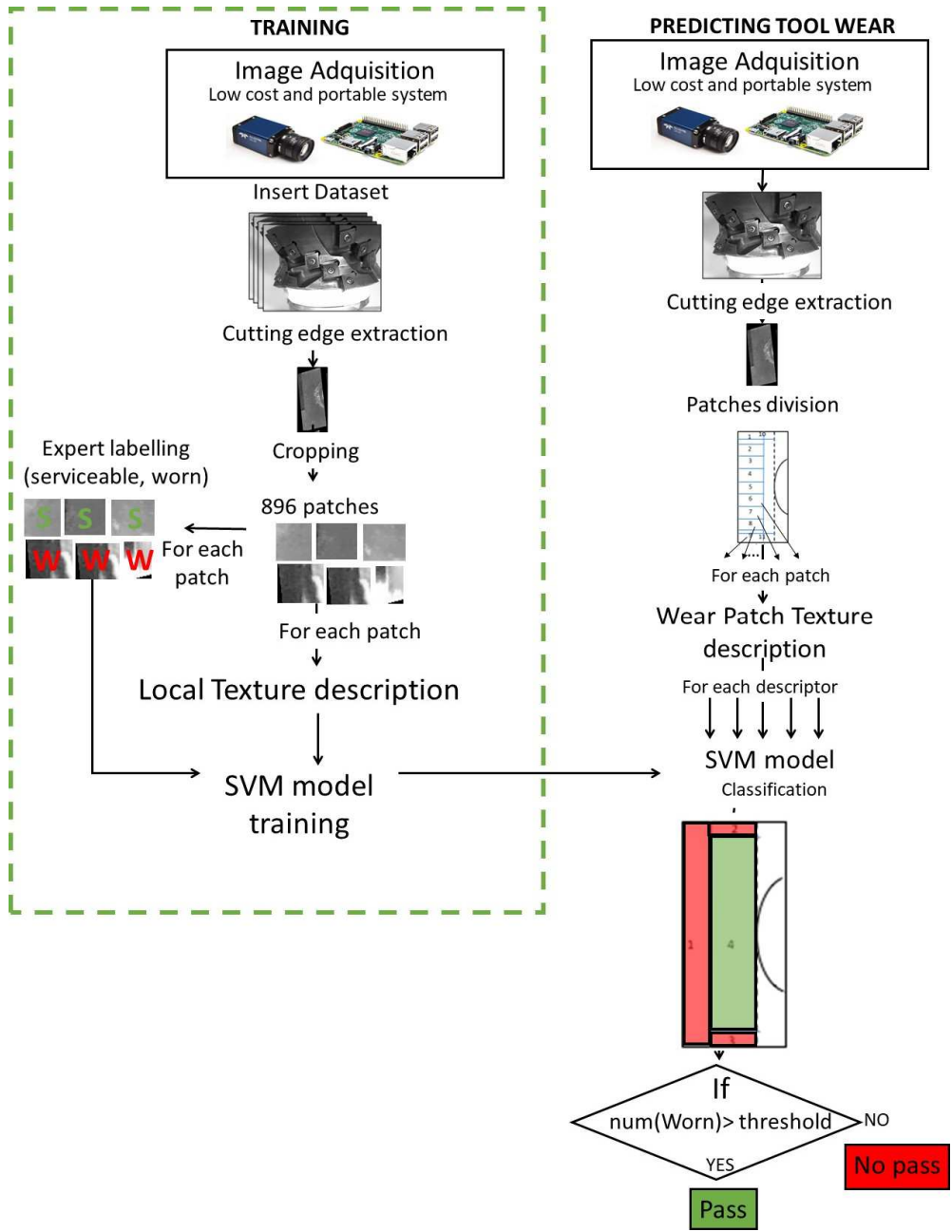

Figure 1: Schema of our proposed system for tool wear monitoring using computer vision.

We refer to this process as portable, meaning that it is a system whose dimensions are small and its cost is low. Such systems, built upon Single Board Computers (SBC) (e.g. Raspberry Pi, Odroid), make possible to implement solutions easily in embedded schemes with low cost and low power consumption, and time performances comparable to those obtained using traditional computers. The main concern with the use of these valuable and flexible alternatives in industrial environments is their sensitivity to certain conditions such as noise or electrical fields. An industrial setting it is likely to be an electrically noisy environment. For this reason, it is important to isolate these systems from noisy power supply, motors or any other sources of interference to meet industrial requirements.

It is also important to highlight that in a real environment the tool wear assessment is carried out while the head tools are resting, which takes between 5 to 30 minutes [34]. The proposed system takes approximately 0.13 seconds to evaluate each insert, so it takes around 4 seconds to assess the entire head tool (including the time to rotate it), which have 30 inserts.

Some works that apply computer vision and portable systems have been proposed recently. For example, Tu et al. introduced a system for honeybee counting using a Raspberry Pi and computer vision techniques [35]. Other applications have also been developed recently with embedded systems and low cost cameras [36, 37] but, to the best of our knowledge, this is the first time that a portable system is applied to wear tool monitoring.

We also introduce a new dataset of 577 images of cutting edges of an edge profile milling machine that is available online[1]. This is the dataset used to evaluate the performance of our approach.

---

[1] http://pitia.unileon.es/varp/node/439

## 3. Materials and Methods

### 3.1. Image acquisition and processing

The tool head has a cylindrical shape and contains 30 inserts arranged in 6 groups of 5 inserts diagonally located along the axial direction of the tool perimeter. We have used a monochrome Genie M1280 1/3 camera with an active resolution of 2592×1944 which mounts an AZURE-2514MM lens with a focal length of 25mm. This camera is handled using a Raspberry Pi which can be easily integrated in any manufacturing system due to its small size. In order to achieve illumination conditions independent to the environment, we have used three LED bar lights (BDBL-R(IR)82/16H). They also provide a higher contrast in the edges of the inserts. We have to highlight that the machine tool where this system was assessed does not use oils, lubricants or other kind of substances on the tools.

An example of the images obtained automatically by the capturing system is shown in Figure 2.

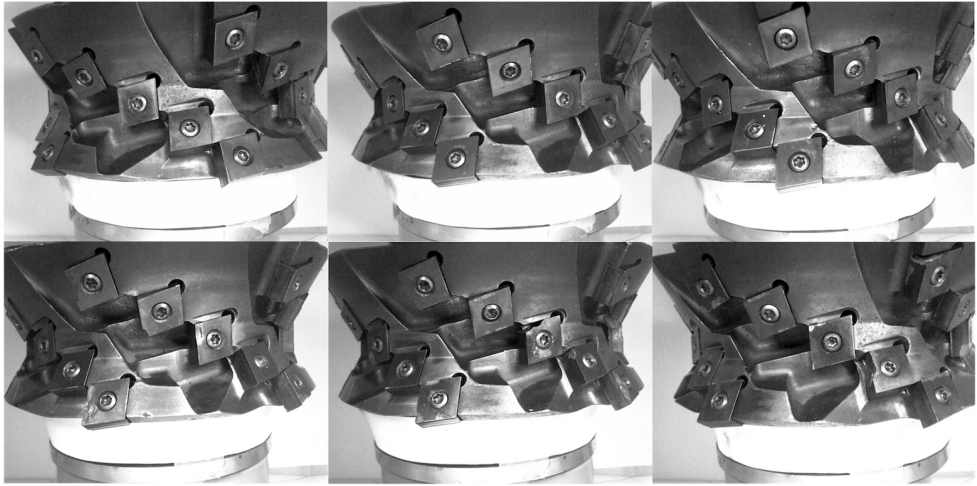

Figure 2: Example of the captured images.

After the image acquisition is finished, the automatic process proposed in [38] to extract the cutting edge is carried out. First, we use the Circular Hough Transform (CHT) [28] to detect the circles with radius between 40 and 80 pixels to detect the screws located in the middle of the inserts. The radius has been fixed as a constant due to the a priori knowledge of the screw size for images of 2592 × 1944. Changing the inserts' geometry or the acquisition system would require tuning this specific parameter using any informal calibration procedure, which can be easily carried out using a few images of the tool head, like the ones shown in Figure 2. Thereafter we apply a Canny edge detector [29] to detect the edges of the inserts. After that, we detect the vertical lines by means of the Standard Hough transform (SHT) [30]. Finally, we extract the cutting edge of the insert from the image taking into account the first vertical line located on the left hand side of the image, which corresponds with the cutting edge due to the spinning direction of the head tool. The final result is shown in Figure 3, where the cutting edges of ten cutting tools are presented. In the first row, serviceable inserts with low tool wear are shown. In the second one, worn inserts with high wear (at different degrees) are presented.

At the end of this process we have obtained 577 images of the cutting edge of the tools. These images were labelled by an expert, so the dataset has 301 disposable edges and 276 worn ones. One expert carried out the labelling process by means of a visual assessment, relying on his previous knowledge and experience. This task is not straightforward as it depends on many parameters like the size of the wear area, its location or how deep it is. Other approaches [11], though, consider several experts opinion in the labelling process, what can improve the performance since it makes the labelling process more reliable.

*3.2. Region Configuration*

We propose a patch-based approach to evaluate the insert wear level. It relies on dividing the cutting edge image into wear patches (WP) with different sizes, shapes and orientations that, later on, are categorized according to their wear (i.e. disposable or serviceable). It is a knowledge-driven patch configuration (i.e. the cutting edge image division) considering both the typical position and shape of the WP. In this section, we present different alternative divisions trying to adapt the patches' shape and position to our region of interest. A graphical representation is shown in Figure 4 and next, we describe each one with more detail.

*3.2.1. Homogeneous Grid Division (HGD)*

In this case the cutting edge image is divided into a 3 x 2 grid. With this configuration, we create patches with the same size in order to give the same importance to all of them in the classification step. The Homogeneous Grid

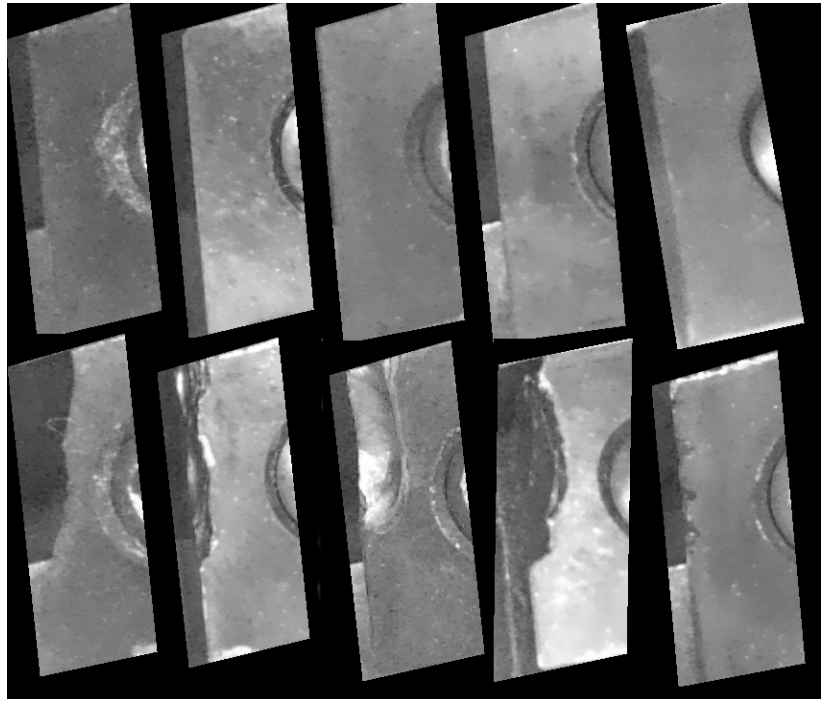

Figure 3: Cutting edge images. In the first row, serviceable edges are shown. In the second one, edges with high wear are displayed.

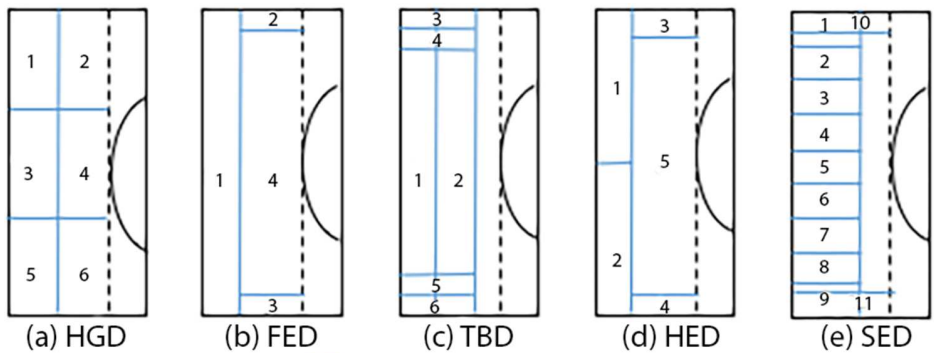

Figure 4: All the different configurations of the patches over the cutting edges.

Division (HGD) is the only configuration that has uniform patch size (See Figure 4(a)).

*3.2.2. Full Edge Division (FED)*

The Full Edge Division (FED) is based on the basic idea of evaluating each cutting edge of the image individually. Figure 4(b) shows a representation of this configuration. For each image, four different patches have been extracted: The main cutting edge patch, two horizontal patches connected with the upper and lower part of the insert, which usually have also some kind of erosion, and finally the interior region of the insert. In this proposal, all the patches are fully independent and includes all the information which appears in the cutting edge image.

*3.2.3. Two Band Division (TBD)*

The TBD alternative, shown in Figure 4(c), extracts six patches per image. The first one, as we did in FED, contains the information of the main cutting edge. Additionally, a patch of the same size as the first one located after the first patch is taken into account in order to capture possible high wear regions located also after the border of the first patch division. We proceed the same way with the horizontal patches, extracting two different patches in the upper zone and two more in the lower one. In this case, the horizontal and vertical patches are overlapped.

*3.2.4. Half Edge Division (HED)*

The Half Edge Division (HED) proposal (see Figure 4(d)), has the same patches configuration as FED, but dividing the vertical patch that contains the edge information into two: The first one containing the information of the main cutting edge located in the top of the insert and the second one with the information of the bottom. With this division, we detect small wear regions that could be misclassified in the first approach due to their specific location in just one part of the vertical patch.

*3.2.5. Small Edge Division (SED)*

Trying to merge some of the ideas used in the other configurations, we propose the Small Edge Division (SED). Same as in the TBD method, we want to extract the information of the high wear region but, instead of using two vertical bands, we use a wider region. Furthermore, we have taken into account the idea of the HED method, which divides the vertical patch in two different regions but in this case we have gone further dividing the vertical region in 9 subregions, as we can see in Figure 4(e). Moreover, we have also evaluated the top and bottom parts of the insert as we did in FED, but starting from the corner of the main edge which causes an overlapping between these patches and the two extreme sub-regions of the vertical one.

*3.3. Texture descriptors*

*3.3.1. Local Binary Pattern (LBP)*

LBP [39] is a gray-scale texture operator that extracts pixel-wise information of an image. For each pixel *c*, LBP takes into account its *P* neighbours within a radius (i.e. distance) *R*. When a neighbour *p* has a grey level (i.e. intensity) value $g_p$ greater than or equal to that of *c*, the value 1 is assigned to it, or 0 otherwise. Thereafter, the LBP for that pixel is calculated by summing up those values multiplied by consecutive powers of 2, as it is stated in Equation (1). Figure 5 depicts an example of the extraction of the LBP of one pixel.

$$LBP_{P,R} = \sum_{p=0}^{P-1} s(g_p - g_c)2^p, \quad s(x) = \begin{cases} 1 & \text{if } x \geq 0 \\ 0 & \text{if } x < 0 \end{cases}, \quad (1)$$

where $g_c$ is the intensity value of central pixel, *P* is the number of neighbours, $g_p$ is the value of its *p*-th neighbour, which lies along the orientation $2\pi p/P$ at a distance *R* (i.e. the radius of the neighbourhood).

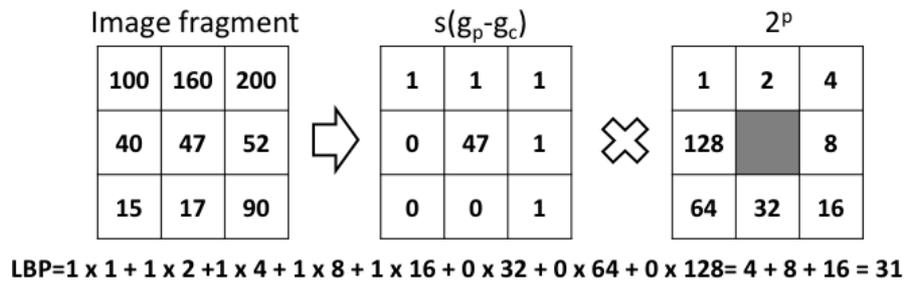

Figure 5: Local Binary Pattern process over one pixel in gray scale level using a neighbourhood of radius 1 and 8 neighbours.

Despite LBP is invariant to monotonic transformations of the gray scale, rotations may yield different LBP values. Therefore, Ojala et al. defined a new formulation of LBP to achieve invariance to rotation by assigning a unique identifier to each rotated LBP [40]. This new definition is stated in Equation (2).

$$LBP_{P,R}^{ri} = \min\{ROR(LBP_{P,R}, i) \mid i = 0, 1, \ldots, P-1\}, \quad (2)$$

where $ROR(x, i)$ is an operator that performs a circular bit-wise right shift $i$ times on the $P$-bit number $x$. For example the rotation invariant patterns of 11001011 and 10010111 have the same value, i.e. 00101111.

*3.3.2. Adaptive Local Binary Pattern (ALBP)*

ALBP is a descriptor based on LBP proposed by Guo et al. [41]. It is motivated by the lack of information about the orientation of conventional LBP. The oriented mean and standard deviations of the local absolute differences $|g_c - g_p|$, $\forall p \in \{0, 1, \ldots, P-1\}$ are taken into account in order to make the matching more robust against local spatial structure changes. Intuitively, the texture classification could be improved by minimising the variations of the mean and standard deviation of the directional differences. Guo et al. introduced a parameter $w_p$ so that the $p$-th directional difference $|g_c - w_p \cdot g_p|$ can be minimised, and defined the following objective function:

$$w_p = \arg\min_w \sum_{i=1}^{N} \sum_{j=1}^{M} |g_c(i,j) - w \cdot g_p(i,j)|^2, \quad (3)$$

where $w_p$ is the weight used to minimise the directional difference, and $N$ and $M$ are the number of rows and columns in the image, respectively. Guo et al. used the least squares estimation method to minimise such objective function and, thus, to obtain the optimum weight parameter vector, i.e. $\overline{w} = [w_0, w_1, \ldots, w_{P-1}]$.

Finally, ALBP is defined as follows:

$$ALBP_{P,R} = \sum_{p=0}^{P-1} s(g_p - w_p \cdot g_c) 2^p, \quad s(x) = \begin{cases} 1 & \text{if } x \geq 0 \\ 0 & \text{if } x < 0 \end{cases}. \quad (4)$$

*3.3.3. Completed Local Binary Pattern (CLBP)*

CLBP was proposed by Guo et al. [42] in order to generalize and complete the classical LBP. In this case, given a central pixel $c$ with intensity $g_c$ and its $P$ neighbours with intensities $g_p$ the local difference vector (i.e. $[d_0, d_1, \ldots, d_{P-1}]$, where $d_p = g_p - g_c$, $\forall p \in 0, 1, \ldots, P-1$), which characterises the image local structure at $g_c$, is taken into account. This local structure is then represented by means of the local difference sign-magnitude transform (LDSMT), which decomposes each $d_p$ into two components: its sign $s_p$ (i.e. 1 if $d_p \geq$ 0 or -1 otherwise), and the magnitude $m_p$ (i.e. $|d_p|$).

Two operators are proposed to model the signs (S) and the magnitudes (M) of the local differences, namely CLBP_S and CLBP_M. The CLBP_S operator is calculated in the same way as LBP histogram (see Equation (1)). CLBP M is defined in Equation (5).

$$CLBP\ M_{P,R} = \sum_{p=0}^{P-1} t(m_p, a) 2^p, \quad t(x, a) = \begin{cases} 1 & \text{if } x \geq a \\ 0 & \text{if } x < a \end{cases} \quad (5)$$

where $a$ is a threshold determined adaptively. In this case we have set it as the mean value of $m_p$.

Finally, since both operators are in binary format, they can be concatenated to form the final CLBP histogram.

*3.3.4. LBP Variance (LBPV)*

LBPV [43], is another proposal made by Guo et al. which combines LBP and a contrast distribution. First, the uniform LBP [39] of the image is calculated. Then, the local variances of the image are used as a weight to adjust the contribution of the LBP code in the histogram calculation The LBPV histogram is computed as:

$$LBPV_{P,R}(k) = \sum_{i=1}^{N} \sum_{j=1}^{M} w(LBP_{P,R}(i,j), k), \quad k \in [0, K], \quad (6)$$

where $k$ is each bin of the histogram, $K$ the maximum value of LBP and $w$ is defined as:

$$w(LBP_{P,R}(i,j), k) = \begin{cases} VAR_{P,R}(i,j), & LBP_{P,R}(i,j) = k \\ 0 & \text{otherwise} \end{cases} \quad (7)$$

In Equation (7) $VAR_{P,R}$ is the rotation invariant measure of the local variance (i.e. the variance of a neighbourhood), defined as:

$$VAR_{P,R} = \frac{1}{P}\sum_{p=0}^{P-1}(g_p - u)^2, \tag{8}$$

where $u = 1/P \sum_{p=0}^{P-1} g_p$.

## 4. Experiments and results

### 4.1. Dataset and experimental setup

There are no publicly available image datasets of tools for milling processes with sufficient quality and large enough to assess texture description methods. For that reason, we created a new dataset comprising 254 images of edge profile cutting heads. Once the the cutting edges are extracted from these (see Section 3.1) the total number of images is 577. In our experiments, the set of cutting edge images has been randomly divided into a training set and a test, which contains 70% and 30% of the whole set, respectively.

In this work we have divided each image of the training set into patches (as per explained in Section 3.2) that later on were labelled as belonging to a worn or functional region. This division has been done manually. The goal in this work is to generate a classification model that assesses individual patches and, based on its predictions, makes a final decision about the tool wear level. Therefore, the manual division avoids possible bad extractions of the patches that could lead to the generation of suboptimal classifiers. After the manual patch extraction we have obtained 896 patches, being 466 of them serviceable and 430 worn.

We have used the descriptors of the patches of the training subset to model a Support Vector Machine (SVM) classifier that uses an intersection kernel. Afterwards, we automatically extract the patches for each image of the test subset using the methods explained in Section 3.2. All these patches were described using the methods based on LBP explained in Section 3.3. After that, each patch was classified using the SVM model in order to determine their labels (i.e. functional or worn) and, finally, we calculated the proportion of deteriorated patches in the image.

We established a threshold parameter in order to determine if an insert is still disposable: if an image contains a number of patches classified as worn higher than the threshold, then the tool edge is considered to be worn. The threshold can vary between one and the number of patches into which the image is divided. The higher the threshold, the stricter it is for the method to label an insert as disposable.

*4.2. Evaluation metrics*

The metrics employed in this paper to test our approach are precision, recall, accuracy and F-score. Considering the worn class as the positive class and the serviceable class as the negative one, a confusion matrix like the shown in Table 1 summarizes the categorization outcomes.

Table 1: Confusion matrix for a binary classification problem

|  |  | Prediction class | |
|---|---|---|---|
|  |  | *worn* (+) | *serviceable* (-) |
| True class | *worn* (+) | tp | fn |
|  | *serviceable* (-) | fp | tn |

Accuracy is computed as the number of successful predictions over the total number of samples as given by (9):

$$Accuracy = \frac{tp + tn}{tp + tn + fp + fn} \quad (9)$$

Precision, also called positive predictive value, is the fraction of inserts classified as worn that are actually worn as indicated by (10).

$$Precision = \frac{tp}{tp + fp} \quad (10)$$

Recall refers to the fraction of worn inserts that are categorized in such class and it can be computed as follows:

$$Recall = \frac{tp}{tp + fn} \quad (11)$$

In this particular problem, the recall metric plays an important role since the cost of misclassifying a worn cutting edge as serviceable is higher than the cost of misclassifying a serviceable cutting edge.

Finally, the F-score metric is defined as the harmonic mean of precision and recall and it can be computed as shown in (12).

$$FScore = \frac{2tp}{2tp + fp + fn} \quad (12)$$

*4.3. Threshold selection*

First of all, we have evaluated all the descriptors with all the different wear region configurations (as per explained in Section 3.2) varying the number of wear patches (WP) that are necessary to consider an insert as worn. In these experiments, we have focused on the recall metric because of the nature of our problem: the impact of misclassifying a worn cutting edge image as serviceable is higher than the other way round. Note that we focus on the evaluation of the cutting edge as a whole with no specific interest in the category each patch falls in.

Figure 6 shows the recall values for all the different experiments using all the assessed threshold values. As expected, in all the cases, the threshold equal to one offers the best recall results with any descriptor and wear region configuration. The higher the threshold value is, the lower the recall achieved with our approach. However, not all the wear region configurations show the same behaviour: the performance using FED, TDB and HED is very dependent on the threshold. For example, using $ALBP_{16,2}$ and FED, the recall varies from 0.8 when the threshold is one to less than 0.2 with threshold equal to four. In contrast, HGD and SED achieve more invariance to the threshold with certain descriptors. For example, the difference with the same thresholds and descriptors is approximately 0.1 in HGD with $ALBP_{16,2}$ on the same interval. This may be due to the higher number of patches that are extracted in these two wear region configurations: the higher the number of divisions, the lower the difference in terms of performance between the thresholds.

It is also interesting that the method that achieves the best recall when the threshold is high is $ALBP_{8,1}$ with a difference of more than 20% with respect to some of the other assessed methods. In summary, for all these reasons, we have fixed the threshold to one.

*4.4. Region wear configuration and description evaluation*

Tables 2, 3, 4, 5 and 6 show the experimental results achieved with region configurations HGD, FED, TBD, HED and SED, respectively, when using a threshold equal to one and all the wear region configuration and descriptor combinations. These results are also depicted in Figure 7. In this case, four different metrics have been taken into account for studying the classification performance: precision, recall, accuracy and F-score. Usually, the F-score is the most representative metric because it takes into account both the precision and recall information. As we can observe, the best F-score is

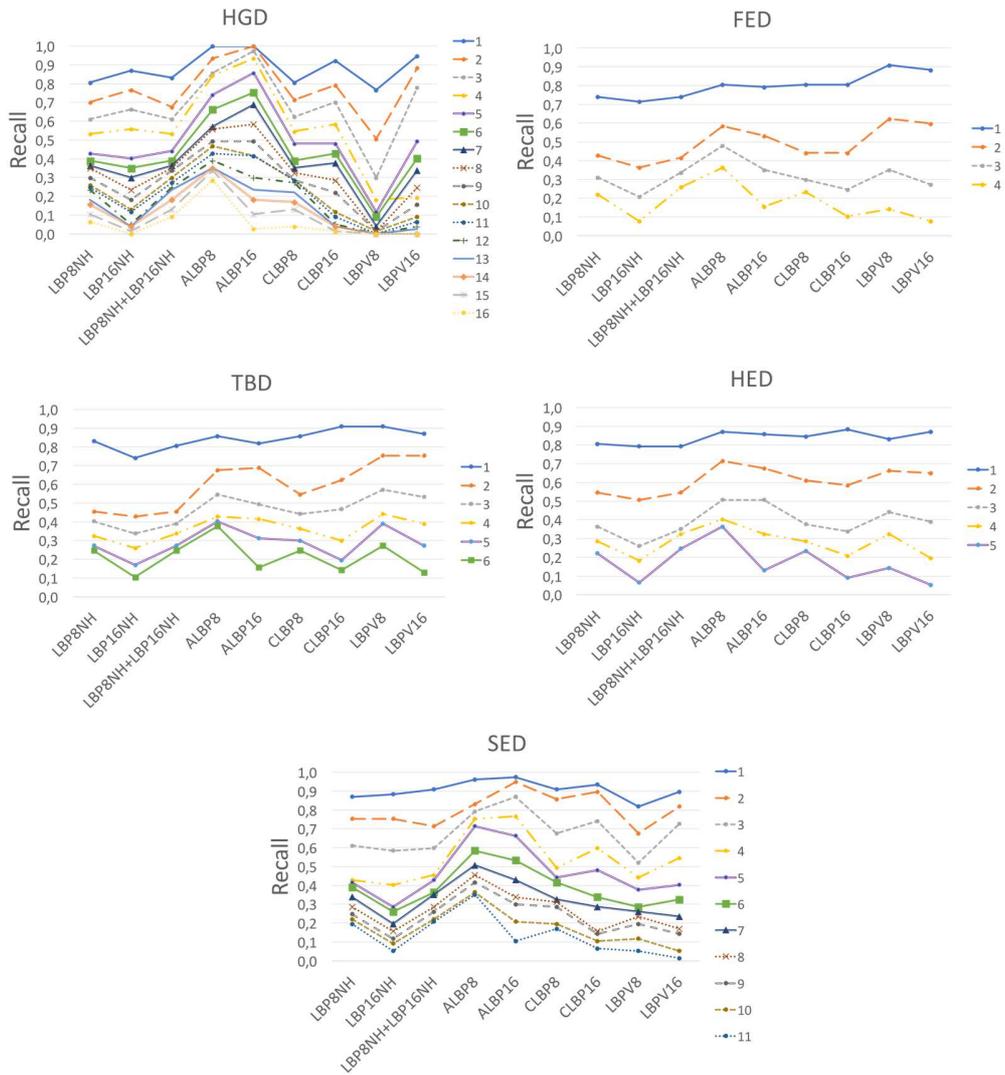

Figure 6: Recall values for all wear region configuration methods using different descriptors and threshold values. For each configuration, each threshold is represented by a different colour, indicated on each legend.

achieved using the TBD wear region configuration and $CLBP_{16,2}$ (0.909) and also using the SED configuration and $LBP_{8,1}+LBP_{16,2}$ (0.903), where the "+" represents the concatenation operator. Both TBD and SED wear region configurations have some overlapping in the patches in the vertical cutting edges (i.e. those in the lower and upper part of the insert), where some wear appears sometimes. This suggests that different possible configurations may consider using overlapped patches in the part of the corners of the cutting edges.

Table 2: Experimental results achieved with region configuration HGD with threshold equal to 1 for the different assessed descriptors.

|   | Precision | Recall | Accuracy | F-score |
|---|---|---|---|---|
| **LBP8NH** | 0.681 | 0.805 | 0.714 | 0.738 |
| **LBP16NH** | 0.684 | 0.870 | 0.734 | 0.766 |
| **LBP8NH + LBP16NH** | 0.800 | 0.831 | 0.812 | 0.815 |
| **ALBP8** | 0.510 | 1.000 | 0.519 | 0.675 |
| **ALBP16** | 0.500 | 1.000 | 0.500 | 0.667 |
| **CLBP8** | 0.681 | 0.805 | 0.714 | 0.738 |
| **CLBP16** | 0.640 | 0.922 | 0.701 | 0.755 |
| **LBPV8** | 0.621 | 0.766 | 0.649 | 0.686 |
| **LBPV16** | 0.541 | 0.948 | 0.571 | 0.689 |

Table 3: Experimental results achieved with region configuration FED with threshold equal to 1 for the different assessed descriptors.

|   | Precision | Recall | Accuracy | F-score |
|---|---|---|---|---|
| **LBP8NH** | 0.919 | 0.740 | 0.838 | 0.820 |
| **LBP16NH** | 0.932 | 0.714 | 0.831 | 0.809 |
| **LBP8NH + LBP16NH** | 0.934 | 0.740 | 0.844 | 0.826 |
| **ALBP8** | 0.667 | 0.805 | 0.701 | 0.729 |
| **ALBP16** | 0.629 | 0.792 | 0.662 | 0.701 |
| **CLBP8** | 0.899 | 0.805 | 0.857 | 0.849 |
| **CLBP16** | 0.912 | 0.805 | 0.864 | 0.855 |
| **LBPV8** | 0.543 | 0.909 | 0.571 | 0.680 |
| **LBPV16** | 0.618 | 0.883 | 0.669 | 0.727 |

It is remarkable that the best results in every region configuration, in terms of F-score and precision have been achieved with either the concatenation of $LBP_{8,1}$ and $LBP_{16,2}$ (in the case of HGD and SED) descriptors or $CLBP_{16,2}$ (in FED, TBD and HED).

Table 4: Experimental results achieved with region configuration TBD with threshold equal to 1 for the different assessed descriptors.

|  | Precision | Recall | Accuracy | F-score |
|---|---|---|---|---|
| **LBP8NH** | 0.914 | 0.831 | 0.877 | 0.871 |
| **LBP16NH** | 0.934 | 0.740 | 0.844 | 0.826 |
| **LBP8NH + LBP16NH** | 0.925 | 0.805 | 0.870 | 0.861 |
| **ALBP8** | 0.635 | 0.857 | 0.682 | 0.729 |
| **ALBP16** | 0.563 | 0.818 | 0.591 | 0.667 |
| **CLBP8** | 0.917 | 0.857 | 0.890 | 0.886 |
| **CLBP16** | 0.909 | 0.909 | 0.909 | 0.909 |
| **LBPV8** | 0.551 | 0.909 | 0.584 | 0.686 |
| **LBPV16** | 0.650 | 0.870 | 0.701 | 0.744 |

Table 5: Experimental results achieved with region configuration HED with threshold equal to 1 for the different assessed descriptors.

|  | Precision | Recall | Accuracy | F-score |
|---|---|---|---|---|
| **LBP8NH** | 0.886 | 0.805 | 0.851 | 0.844 |
| **LBP16NH** | 0.871 | 0.792 | 0.838 | 0.830 |
| **LBP8NH + LBP16NH** | 0.910 | 0.792 | 0.857 | 0.847 |
| **ALBP8** | 0.638 | 0.870 | 0.688 | 0.736 |
| **ALBP16** | 0.569 | 0.857 | 0.604 | 0.684 |
| **CLBP8** | 0.844 | 0.844 | 0.844 | 0.844 |
| **CLBP16** | 0.861 | 0.883 | 0.870 | 0.872 |
| **LBPV8** | 0.533 | 0.831 | 0.552 | 0.650 |
| **LBPV16** | 0.644 | 0.870 | 0.695 | 0.740 |

Table 6: Experimental results achieved with region configuration SED with threshold equal to 1 for the different assessed descriptors.

|  | Precision | Recall | Accuracy | F-score |
|---|---|---|---|---|
| **LBP8NH** | 0.817 | 0.870 | 0.838 | 0.843 |
| **LBP16NH** | 0.895 | 0.883 | 0.890 | 0.889 |
| **LBP8NH + LBP16NH** | 0.897 | 0.909 | 0.903 | 0.903 |
| **ALBP8** | 0.552 | 0.961 | 0.591 | 0.701 |
| **ALBP16** | 0.532 | 0.974 | 0.558 | 0.688 |
| **CLBP8** | 0.795 | 0.909 | 0.838 | 0.848 |
| **CLBP16** | 0.818 | 0.935 | 0.864 | 0.873 |
| **LBPV8** | 0.685 | 0.818 | 0.721 | 0.746 |
| **LBPV16** | 0.633 | 0.896 | 0.688 | 0.742 |

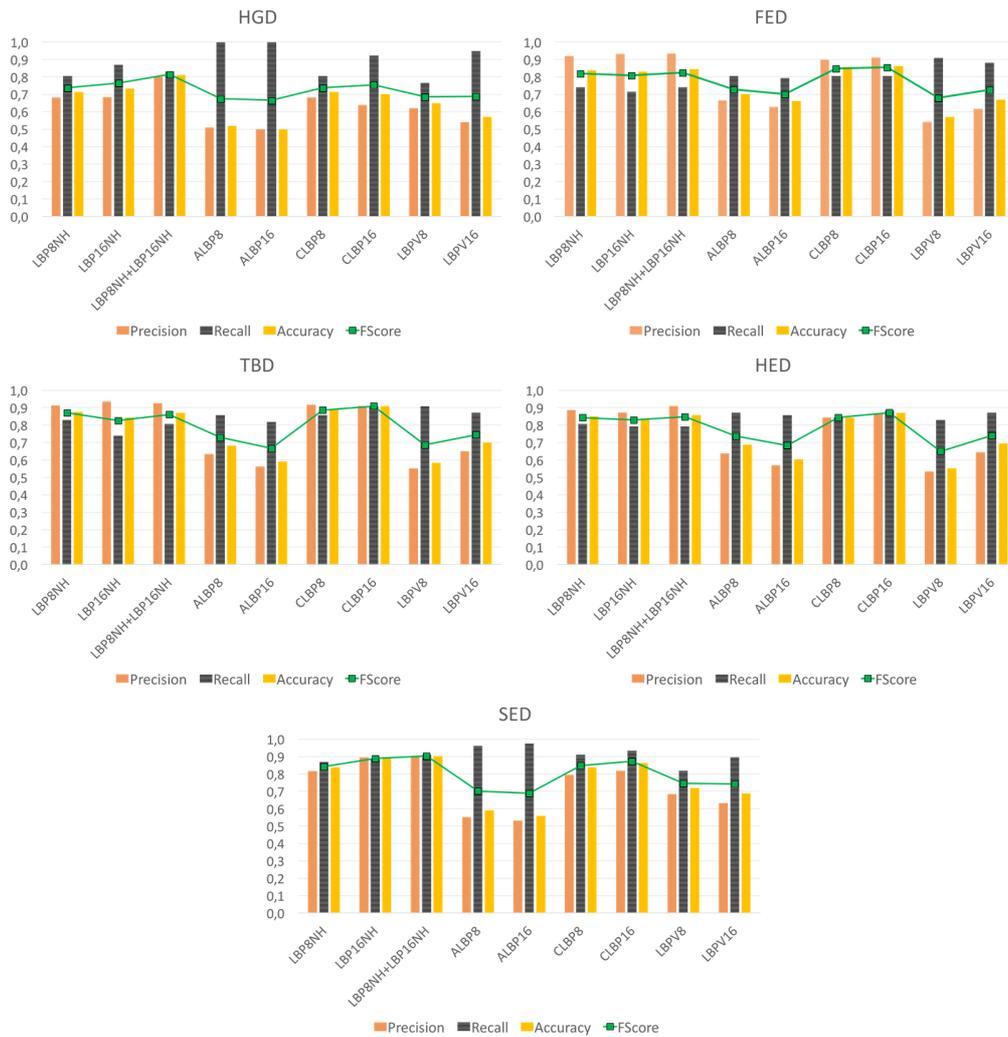

Figure 7: Results of accuracy, precision, recall and F-score with all the methods and descriptors for the selected threshold value equal to one.

Another interesting consideration is that although HGD and SED achieve high values of recall using $ALBP_{8,1}$ and $ALBP_{16,2}$, the precision is too low to consider them as a suitable solution.

Taking into account the information depicted in Figure 7 and Tables 2 - 6, we can state that the best methods are either the combination composed by the TBD wear region schema and the $CLBP_{16,2}$ descriptor, due to its high recall (0.909), F-score (0.909) and accuracy of 90.91%, or SED and $LBP_{8,1}+LBP_{16,2}$ which achieves a recall of 0.909, F-score of 0.903 and high accuracy (90.26%). However, as it is depicted in Figure 6, SED provides the most stable results in terms of threshold variation which allows the expert more flexibility when varying the threshold for cases when the precision would be more important than usual. This is due to the higher number of patches in the SED region configuration (i.e. 11) in comparison with TBD (i.e. 6). Figure 8 depicts some edges right and wrongly classified as worn and serviceable using this recommended configuration (i.e. SED with $LBP_{8,1}+LBP_{16,2}$).

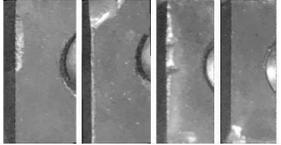

Figure 8: Confusion matrix showing edges well and bad classified as worn or serviceable. Rows represent the true class (i.e. worn or serviceable) and the columns represent the predictions made by our proposal.

We also compared our proposal with other classical descriptors, i.e. Bag of Contour Fragments (BCF) [44], Histogram of Oriented Gradients (HOG) [45], Shape Context (SC) [46] and also with the descriptor B-ORCHIZ proposed by García-Ordás et al. in [19], extracted from a similar dataset. We show the results in Table 7, where we can see that all these state-of-the-art descriptors are outperformed by our proposal. Apart from ours, the best one was B-ORCHIZ, with an accuracy of 87.06%, which was a descriptor composed by 100 features which combined local and global information of the shape of the cutting edge wear region. The classification was carried out with a Support Vector Machine (SVM) with Intersection kernel. Its main drawback was that it required a manual segmentation of the worn region of each insert, which is very time consuming (thus, forcing the process to be off-line) and is prone to errors due to inaccurate segmentations. It was, therefore, a less interesting solution to the manufacturing community.

Table 7: Classification accuracy in % of the state-of-the-art descriptors Bag of contour fragments (BCF) [44], Histogram of oriented gradients (HOG) [45], Shape Context (SC) [46], B-ORCHIZ [19] and our proposed method using the SED configuration and the texture descriptors $LBP_{8,1}+LBP_{16,2}$.

| Descriptor | Accuracy (%) |
|---|---|
| SC [46] | 54.58 |
| BCF [44] | 76.76 |
| HOG [45] | 76.80 |
| B-ORCHIZ [19] | 87.06 |
| **SED with $LBP_{8,1}+LBP_{16,2}$** | **90.26** |

## 5. Conclusion

In conclusion, we have proposed a new method to detect the state of cutting edges (i.e. serviceable or disposable) of inserts in edge profiles milling processes. Our proposal is an online method based on low cost devices such as the Raspberry Pi and a small monochrome camera (Genie M1280).

The proposed method divides the cutting edges into several sub-regions (i.e. the wear patches (WP)). Each WP is described using texture descriptors based on Local Binary Pattern (LBP) and classified by means of a Support Vector Machine (SVM) as worn or serviceable. Finally, the decision about whether a cutting edge is serviceable or disposable is based on the number of WP classified as worn.

We have presented several configurations to divide the cutting edge images: Homogeneous grid division (HGD), full edge division (FED), two band division (TBD), half edge division (HED) and small edge division (SED). We have demonstrated that when at least one of these WP is classified as worn, it is more likely that the insert is disposable.

Using a concatenation of the descriptors $LBP_{8,1}$ and $LBP_{16,2}$ we have achieved an F-score of 0.903 and an accuracy 90.26% dividing the cutting edge following the SED configuration, what outperforms previous approaches [19]. Unlike [19], our method does not require segmentation of the worn region, which is very time consuming (thus, forcing the process to be off-line) and subject to segmentation errors. Therefore, the approach presented in this work achieves high accuracy and does not require a segmentation stage, what makes feasible to deploy a fully automatic online method to characterise the tool inserts.

The proposed methodology has been assessed on a specific type of inserts, whereas a wide variety of situations may be found in real industrial environments, such as different materials being machined or inserts with different geometry. Anyway, the proposed method can be extended to any other tool or material. For example, transferring this proposal to a production environment where other materials are machined would require the collection of a representative set of images of inserts labelled by experts in terms of the wear degree and the subsequent training of the categorization module. When the tool geometry changes, the image preprocessing module should be adapted accordingly, e.g. to be able to extract successfully the inserts. The proposals for the division of the cutting edge in patches and texture features proposed (i.e. $LBP_{8,1}+LBP_{16,2}$ or $CLBP_{16,2}$) can thereafter be used to create a model which automatically characterizes the wear level of new unseen tools. Anyway, in case any other region configuration needs to be used, we recommend the use of a region configuration with an overlapping of patches in the main and secondary edges, where some wear may appear, and a high number of patches.

**Acknowledgements**

This work has been supported by the research project with ref. DPI2012-36166 from the Spanish Ministry of Economy and Competitiveness and the PIRTU program of the Regional Government of Castilla y León.

# 6. References

[1] P. Hosseini and P. Radziszewski, "Combined study of wear and abrasive fragmentation using steel wheel abrasion test," *Wear*, vol. 271, no. 5, pp. 689 – 696, 2011.

[2] X.-m. Zhang and W.-p. Chen, "Review on corrosion-wear resistance performance of materials in molten aluminum and its alloys," *Transactions of Nonferrous Metals Society of China*, vol. 25, no. 6, pp. 1715–1731, 2015.

[3] F. Cus and U. Zuperl, "Real-time cutting tool condition monitoring in milling," *Strojniski vestnik-Journal of Mechanical Engineering*, vol. 57, no. 2, pp. 142–150, 2011.

[4] M. Nouri, B. K. Fussell, B. L. Ziniti, and E. Linder, "Real-time tool wear monitoring in milling using a cutting condition independent method," *International Journal of Machine Tools and Manufacture*, vol. 89, pp. 1 – 13, 2015.

[5] T.-I. Liu and B. Jolley, "Tool condition monitoring (tcm) using neural networks," *The International Journal of Advanced Manufacturing Technology*, vol. 78, no. 9-12, pp. 1999–2007, 2015.

[6] D. Bajic, L. Celent, and S. Jozic, "Modeling of the influence of cutting parameters on the surface roughness, tool wear and cutting force in face milling in off-line process control," *Journal of Mechanical Engineering*, vol. 58, no. 11, pp. 673–682, 2012.

[7] S. Painuli, M. Elangovan, and V. Sugumaran, "Tool condition monitoring using k-star algorithm," *Expert Systems with Applications*, vol. 41, no. 6, pp. 2638 – 2643, 2014.

[8] W. Rmili, A. Ouahabi, R. Serra, and R. Leroy, "An automatic system based on vibratory analysis for cutting tool wear monitoring," *Measurement*, vol. 77, pp. 117 – 123, 2016.

[9] T. S. Reddy, C. E. Reddy, and S. Prabhavathi, "Belief network-based acoustic emission analysis for real time monitoring in cim environment," *Journal of Engineering and Technology Research*, vol. 3, no. 4, pp. 127–132, 2011.

[10] M. García-Ordás, E. Alegre, V. González-Castro, and D. García-Ordás, "aZIBO: A New Descriptor Based in Shape Moments and Rotational Invariant Features," in *Pattern Recognition (ICPR), 2014 22nd International Conference on*, Aug 2014, pp. 2395–2400.

[11] S. Soleimani, J. Sukumaran, A. Kumcu, P. D. Baets, and W. Philips, "Quantifying abrasion and micro-pits in polymer wear using image processing techniques," *Wear*, vol. 319, no. 1, pp. 123 – 137, 2014.

[12] M. Castejón, E. Alegre, J. Barreiro, and L. Hernández, "On-line tool wear monitoring using geometric descriptors from digital images," *International Journal of Machine Tools and Manufacture*, vol. 47, no. 12, pp. 1847–1853, 2007.

[13] S. Dutta, S. K. Pal, and R. Sen, "Progressive tool flank wear monitoring by applying discrete wavelet transform on turned surface images," *Measurement*, vol. 77, pp. 388 – 401, 2016.

[14] A. Antić, B. Popović, L. Krstanović, R. Obradović, and M. Milosević, "Novel texture-based descriptors for tool wear condition monitoring," *Mechanical Systems and Signal Processing*, vol. 98, pp. 1 – 15, 2018.

[15] T. Miko-lajczyk, K. Nowicki, A. K-lodowski, and D. Pimenov, "Neural network approach for automatic image analysis of cutting edge wear," *Mechanical Systems and Signal Processing*, vol. 88, pp. 100 – 110, 2017.

[16] J. Cervantes, J. Taltempa, F. Garcia-Lamont, J. S. Ruiz-Castilla, A. Y. Rendon, and L. D. Jalilia, "Comparative analysis of the techniques used in a recognition system of plant leaves," *RIAI Revista Iberoamericana de Automática e Informática Industrial*, vol. 14, no. 1, pp. 104 – 114, 2017.

[17] T. Miko-lajczyk, K. Nowicki, A. Bustillo, and D. Y. Pimenov, "Predicting tool life in turning operations using neural networks and image processing," *Mechanical Systems and Signal Processing*, vol. 104, pp. 503 – 513, 2018.

[18] D. M. D'Addona, A. M. M. S. Ullah, and D. Matarazzo, "Tool-wear prediction and pattern-recognition using artificial neural network and dna-based computing," *Journal of Intelligent Manufacturing*, vol. 28, no. 6, pp. 1285–1301, Aug 2017.

[19] M. T. García-Ordás, E. Alegre, V. González-Castro, and R. Alaiz-Rodríguez, "A computer vision approach to analyze and classify tool wear level in milling processes using shape descriptors and machine learning techniques," *The International Journal of Advanced Manufacturing Technology*, vol. 90, no. 5-8, pp. 1947–1961, 2017.

[20] F. M. Anuar, R. Setchi, and Y. kun Lai, "Trademark image retrieval using an integrated shape descriptor," *Expert Systems with Applications*, vol. 40, no. 1, pp. 105 – 121, 2013.

[21] A. Datta, S. Dutta, S. Pal, and R. Sen, "Progressive cutting tool wear detection from machined surface images using voronoi tessellation method," *Journal of Materials Processing Technology*, vol. 213, no. 12, pp. 2339 – 2349, 2013.

[22] A. Kassim, M. Mannan, and Z. Mian, "Texture analysis methods for tool condition monitoring," *Image and Vision Computing*, vol. 25, no. 7, pp. 1080 – 1090, 2007, computer Vision Applications.

[23] E. Gadelmawla, "A vision system for surface roughness characterization using the gray level co-occurrence matrix," *NDT & E International (Independent Nondestructive Testing and Evaluation)*, vol. 37, no. 7, pp. 577 – 588, 2004.

[24] S. Dutta, A. Datta, N. D. Chakladar, S. Pal, S. Mukhopadhyay, and R. Sen, "Detection of tool condition from the turned surface images using an accurate grey level co-occurrence technique," *Precision Engineering*, vol. 36, no. 3, pp. 458 – 466, 2012.

[25] R. S. Samik Dutta, Surjya K. Pal, "On-machine tool prediction of flank wear from machined surface images using texture analyses and support vector regression," *Precision Engineering*, vol. 43, pp. 34 – 42, 2016.

[26] L. Fernández-Robles, G. Azzopardi, E. Alegre, and N. Petkov, "Machine-vision-based identification of broken inserts in edge profile milling heads," *Robotics and Computer-Integrated Manufacturing*, vol. 44, pp. 276 – 283, 2017.

[27] K. Zhu and X. Yu, "The monitoring of micro milling tool wear conditions by wear area estimation," *Mechanical Systems and Signal Processing*, vol. 93, pp. 80 – 91, 2017.

[28] R. O. Duda and P. E. Hart, "Use of the Hough Transformation to Detect Lines and Curves in Pictures," *Commun. ACM*, vol. 15, no. 1, pp. 11–15, Jan. 1972.

[29] J. Canny, "A computational approach to edge detection," *IEEE Transactions on Pattern Analysis and Machine Intelligence*, vol. PAMI-8, no. 6, pp. 679–698, Nov 1986.

[30] P. Hough, "Method and means for recognizing complex patterns," 12 1962, uS Patent 3,069,654.

[31] X. Li, Q. Ruan, Y. Jin, G. An, and R. Zhao, "Fully automatic 3D facial expression recognition using polytypic multi-block local binary patterns," *Signal Processing*, vol. 108, pp. 297 – 308, 2015.

[32] A. Halidou, X. You, M. Hamidine, R. A. Etoundi, L. H. Diakite, and Souleimanou, "Fast pedestrian detection based on region of interest and multi-block local binary pattern descriptors," *Computers & Electrical Engineering*, vol. 40, no. 8, pp. 375 – 389, 2014.

[33] V.-D. Hoang, M.-H. Le, and K.-H. Jo, "Hybrid cascade boosting machine using variant scale blocks based HOG features for pedestrian detection," *Neurocomputing*, vol. 135, pp. 357 – 366, 2014.

[34] L. Fernández-Robles, G. Azzopardi, E. Alegre, and N. Petkov, *Cutting Edge Localisation in an Edge Profile Milling Head.* Cham: Springer International Publishing, 2015, ch. Computer Analysis of Images and Patterns: 16th International Conference, CAIP 2015, Valletta, Malta, September 2-4, 2015, Proceedings, Part II, pp. 336–347.

[35] G. J. Tu, M. K. Hansen, P. Kryger, and P. Ahrendt, "Automatic behaviour analysis system for honeybees using computer vision," *Computers and Electronics in Agriculture*, vol. 122, pp. 10 – 18, 2016.

[36] E. Shakshuki, K. Hammoudi, H. Benhabiles, M. Kasraoui, N. Ajam, F. Dornaika, K. Radhakrishnan, K. Bandi, Q. Cai, and S. Liu, "Developing vision-based and cooperative vehicular embedded systems for enhancing road monitoring services," *Procedia Computer Science*, vol. 52, pp. 389 – 395, 2015, the 6th International Conference on Ambient Systems, Networks and Technologies (ANT-2015), the 5th International Conference on Sustainable Energy Information Technology (SEIT-2015).

[37] L. F. Cambuim, R. M. Macieira, F. M. Neto, E. Barros, T. B. Ludermir, and C. Zanchettin, "An efficient static gesture recognizer embedded system based on ELM pattern recognition algorithm," *Journal of Systems Architecture*, vol. 68, pp. 1 – 16, 2016.

[38] L. Fernández-Robles, G. Azzopardi, E. Alegre, and N. Petkov, "Cutting edge localisation in an edge profile milling head," in *International Conference on Computer Analysis of Images and Patterns*. Springer, 2015, pp. 336–347.

[39] T. Ojala, M. Pietikainen, and D. Harwood, "A comparative study of texture measures with classification based on featured distributions," *Pattern Recognition*, vol. 29, no. 1, pp. 51 – 59, 1996.

[40] T. Ojala, M. Pietikäinen, and T. Mäenpää, "Multiresolution gray-scale and rotation invariant texture classification with local binary patterns," *IEEE Trans. Pattern Anal. Mach. Intell.*, vol. 24, no. 7, pp. 971–987, Jul. 2002.

[41] Z. Guo, L. Zhang, D. Zhang, and S. Zhang, "Rotation invariant texture classification using adaptive lbp with directional statistical features," in *Image Processing (ICIP), 17th IEEE International Conference on*. IEEE, 2010, pp. 285–288.

[42] Z. Guo, L. Zhang, and D. Zhang, "A completed modeling of local binary pattern operator for texture classification," *IEEE Transactions on Image Processing*, vol. 19, no. 6, pp. 1657–1663, 2010.

[43] ——, "Rotation invariant texture classification using LBP variance (LBPV) with global matching," *Pattern recognition*, vol. 43, no. 3, pp. 706–719, 2010.

[44] X. Wang, B. Feng, X. Bai, W. Liu, and L. J. Latecki, "Bag of contour fragments for robust shape classification," *Pattern Recognition*, vol. 47, no. 6, pp. 2116 – 2125, 2014.

[45] N. Dalal and B. Triggs, "Histograms of oriented gradients for human detection," in *2005 IEEE Computer Society Conference on Computer Vision and Pattern Recognition (CVPR'05)*, vol. 1, June 2005, pp. 886–893 vol. 1.

[46] S. Belongie, J. Malik, and J. Puzicha, "Shape matching and object recognition using shape contexts," *IEEE Transactions on Pattern Analysis and Machine Intelligence*, vol. 24, no. 4, pp. 509–522, Apr 2002.